\documentclass[10pt, a4paper]{article}
\usepackage{tcolorbox}
\usepackage[final]{lrec2026} 
\title{Analysing Personal Attacks in U.S. Presidential Debates}

\name{Ruban Goyal$^{1}$, Rohitash Chandra$^{2}$, Sonit Singh$^{1}$} 

\address{$^{1}$School of Computer Science and Engineering, University of New South Wales, Sydney, Australia \\
         $^{2}$School of Mathematics and Statistics, University of New South Wales, Sydney, Australia \\
         ruban.goyal@unswalumni.com, rohitash.chandra@unsw.edu.au, sonit.singh@unsw.edu.au \\}

\abstract{
Personal attacks have become a notable feature of U.S. presidential debates and play an important role in shaping public perception during elections. Detecting such attacks can improve transparency in political discourse and provide insights for journalists, analysts and the public. Advances in deep learning and transformer-based models, particularly BERT and large language models (LLMs) have created new opportunities for automated detection of harmful language. Motivated by these developments, we present a framework for analysing personal attacks in U.S. presidential debates. Our work involves manual annotation of debate transcripts across the 2016, 2020 and 2024 election cycles, followed by statistical and language-model based analysis. We investigate the potential of fine-tuned transformer models alongside general-purpose LLMs to detect personal attacks in formal political speech. This study demonstrates how task-specific adaptation of modern language models can contribute to a deeper understanding of political communication.
\\ \newline  }

\begin{document}

\maketitleabstract

\section{Introduction}

Personal attacks during political debates have gained increasing attention in recent years, especially in the context of U.S. presidential elections~\cite{Egelhofer, MACAGNO202267}. These attacks shape public opinion and often deepen political division~\cite{RODRIGOGINES}. Automatically detecting such attacks is important as it can improve the quality of political discourse and provide fairer grounds for media and public analysis~\cite{Mina_Momeni}.

Personal attacks in high-profile debates reduce the quality of discussions and can distort how the public interprets political messages. Manual detection is possible but it is time-consuming and often inconsistent across annotators. The growth of artificial intelligence provides an opportunity to automate this process. Advances in deep learning allow faster, scalable, and more consistent analysis of political discourse. This study was motivated by the goal of exploring how fine-tuned and general-purpose AI models can contribute to greater transparency and accountability in politics.

This paper explores how modern language models can be applied to detect personal attacks in political debates. The study considers fine-tuned transformer models such as BERT~\cite{devlin-etal-2019-bert} as well as large language models (LLMs) including ChatGPT~\cite{Brown2020gpt3,openai2023gpt4}, Claude~\cite{claude3}, Gemini~\cite{geminiteam2023gemini}, Grok, and DeepSeek~\cite{deepseekai2024}. It also tests the impact of task-specific training by fine-tuning the Meta-LLaMA-3-3B-Instruct~\cite{touvron2023llama} model with LoRA~\cite{LoRA}. The overall aim is to assess and improve automated detection methods so that political communication becomes more transparent and respectful.

This research addresses a real-world challenge in the political domain. It combines manual annotations with predictions from fine-tuned BERT models and large language models to create a layered analysis of personal attacks in debates. The study focuses on U.S. presidential debates from 2016, 2020, and 2024, providing insights into how personal attacks have evolved over time. The findings highlight the strengths and weaknesses of different AI approaches in understanding political language. These results can help journalists, analysts, and the public critically engage with political speech. The paper made the following four contributions:

\begin{itemize}
    \item \textit{Data collection and analysis}: We collected debate transcripts from the 2016, 2020, and 2024 United States (U.S.) presidential elections. We also defined the annotations guidelines and manually annotated the curated dataset with labels indicating whether there is a personal attack or no in the transcript. 
    
    \item \textit{Evaluation of Existing Models}: We applied various pre-trained models that were fine-tuned on public hate speech datasets to the annotated transcripts and assessed how well these models adapt to political debate language before task-specific fine-tuning.
    
    \item \textit{Fine-tuning BERT model}: We fine-tuned Bidirectional Encoder Representations from Transformers (BERT) model on manually annotated transcripts to assess whether fine-tuning helps in detecting personal attacks. 
    
    \item \textit{Comparison with SOTA LLMs}: We also established an evaluation framework to compare fine-tuned BERT models and large language models (LLMs), including ChatGPT, Claude, Gemini, Grok, and fine-tuned LLaMA. 
\end{itemize}

This paper is structured as follows: Section 2 provides technical background on natural language processing and models used in this
study. Section 3 reviews related work and existing datasets. Section 4 presents the implementation and results from fine-tuned BERT models and LLM predictions. Finally, section 5 concludes and discusses possible future directions.

\section{Background and Related Work}~\label{sec:rl}

Natural Language Processing (NLP) is a field of Artificial Intelligence that focuses on enabling computers to understand, interpret and respond to human language. It consists of various tasks ranging from sentiment analysis to more complex tasks like segmentation or entity recognition. NLP has evolved over decades, from early rule-based systems to modern neural network-based models that excel in language understanding and generation~\cite{khurana2023natural}.

\subsection{Statistical models}
A statistical model is mathematical framework that uses data patterns to make predictions and categorise information. N-grams are a fundamental concept in NLP. It represents contiguous sequences of 'n' items, typically words within a text. The value of n determines the length of these sequences. \emph{Bigrams} refers to sequences of two adjacent words. For example, in the sentence "The dog barks", the bigrams would be "The dog" and "dog barks"~\cite{cavnar1994n}. \emph{Trigrams:} It refers to sequences of three adjacent words. In the same sentence, the trigram would be "The dog barks"~\cite{cavnar1994n}. N-grams are used for frequency-based analysis where they help identify commonly occurring word pairs or triplets in a text. They provide essential insights into the structure of a document and are frequently used as a precursor to more advanced research as it helps to establish key insights.

\subsection{Transformer-based models}
Transformer-based models have revolutionised natural language processing (NLP) by leveraging self-attention mechanisms which allow the model to weigh the relevance of each word in sentence irrespective of its position. Unlike Recurrent Neural Networks (RNNs), transformers enable parallel processing which makes them more efficient for large scale tasks. The transformer model is foundation for many state of the art NLP models.

\begin{figure}[]
    \centering
    \includegraphics[scale=0.35]{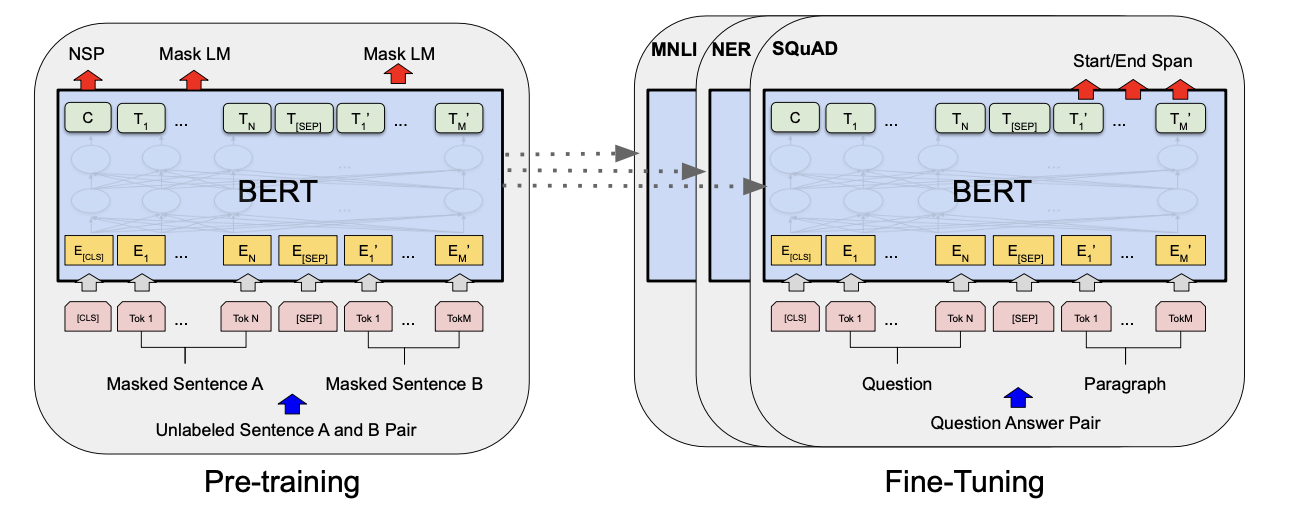}
    \caption{BERT Architecture~\cite{DBLP:journals}}
    \label{fig:BERT}
\end{figure}

Bidirectional Encoder Representations from Transformer (BERT) is a transformer-based model designed to improve language understanding by utilising bidirectional training. This means that BERT processes input text by looking at both the left and right context of each word. Hence, BERT is able to achieve better understanding of language as compared to other models that process text in a single direction~\cite{DBLP:journals}. Figure~\ref{fig:BERT} provides an overview of the BERT architecture. The BERT model has better understanding of context because of its bidirectional training. Also, BERT is pre-trained on large datasets and can be fine-tuned for specific tasks with minimal additional training data~\cite{DBLP:journals}.

\subsection{Large Language Models}

The Large Language Models (LLMs) are AI models designed to understand and generate human language. These models are built using a combination of neural networks and machine learning algorithms. LLMs are trained on massive amounts of text data such as Wikipedia, news articles and social media posts. This allows them to learn the patterns and relationships that exist within language and use this knowledge to generate responses, complete tasks and even write coherent pieces of text~\cite{ray2023chatgpt}. Various popular LLMs include, ChatGPT~\cite{ray2023chatgpt}, Gemini~\cite{imran2024google}, Claude, Grok, DeepSeek, and LLaMa. 

\subsection{Related Work}
There has been few works related to hate speech or personal attack detection in the literature. In~\cite{BidenvsTrump}, BERT is used for sentiment analysis in US General Elections to predict the outcome of 2020 US Presidential Election between Biden and Trump. The study explores how social media platforms like Twitter can be valuable in understanding voter behaviour.
The framework helped identify which states leaned toward Biden or Trump and highlighted key swing states. The HateBERT~\cite{HateBERT} was fine tuned to specialise in abusive language detection by retraining BERT on offensive and abusive comments. Hence, It offers a good prospect to be tested to detect personal attacks or insults in US presidential debates. The HateBERT model outperformed the general purpose BERT model in various datasets such as OffensEval 2019, AbusEval and HatEval. In \cite{gladwin2022toxicbert}, authors investigated the effectiveness of BERT and Support Vector Machine (SVM) in classifying toxic comments on social media platforms. The authors used the \emph{Toxic Comment Classification Challenge} dataset from the Kaggle. The dataset comprises over 150,000 comments sourced primarily from Wikipedia Talk pages and has six categories such as \emph{Toxic}, \emph{Severe-Toxic}, \emph{Obscene}, \emph{Threat}, \emph{Insult} and \emph{Identity-hate}. The BERT outperformed SVM across all toxic comment categories. The result emphasises the advantage of transformer-based models for tasks such as toxic language detection where contextual understanding is critical. It also implies that BERT is able to capture both the explicit and implicit meaning within toxic comments more effectively than SVM. In \cite{fluit2020polarization}, authors manually identified ad hominem (personal attack) and their effects on political polarisation but the approach lacked scalability due to manual process. \cite{Singh:2025:HP_BERT} highlighted BERT's ability to understand complex relationship in text and provides foundation for exploring personal attack detection in political debates.

\section{Methodology}~\label{sec:methodology}

This section outlines the approach taken to detect personal attacks in US presidential debates using transformer-based models and large language models.

\subsection{Dataset collection, annotation and analysis}

Debate transcripts from the 2016, 2020, and 2024 US presidential election cycles were collected as the foundation for this study. The transcripts were sourced from the American Presidency Project \cite{APP}.

The raw text files were converted into structured data using a Python script. Each transcript was read line by line to preserve sequence. Speakers were identified by parsing uppercase names followed by colons, and their utterances were stored in dictionaries with two fields: \texttt{Speaker} and \texttt{Text}. These dictionaries were transformed into a DataFrame and exported in a csv format. A validation step was included to ensure that the converted data matched the original transcripts.

To build a labelled dataset, each line of dialogue was manually annotated using a set of ten guiding questions designed to identify personal attacks, as given below.
These questions were designed to identify whether a statement constitutes a personal attack:

\begin{enumerate}
    \item Is the statement directly aimed at the opponent's character or integrity?
    \item Does it target personal traits rather than political positions or policies?
    \item Is it based on ridicule, sarcasm or mockery?
    \item Does it involve name-calling or derogatory language?
    \item Is it an accusation of misconduct or criminal behaviour?
    \item Is it emotionally charged rather than fact-based?
    \item Does it question the opponent’s intelligence, competence or mental fitness?
    \item Is it focused on the opponent's family or personal life?
    \item Is it about the opponent's physical appearance or health?
    \item Does it use guilt by association to discredit the opponent?
\end{enumerate}

Examples of these criteria included whether a statement targeted the opponent’s character, used mockery, questioned intelligence, or referred to personal life or appearance. Each line was assigned a binary label: \texttt{1} for personal attack and \texttt{0} for non-attack. These annotations were recorded in the csv file. Statements satisfying either of the ten conditions were marked as personal attacks.

\textbf{Example:}
\begin{itemize}
    \item Non-Attack: ``OK, Vice President Biden, your response please?" (Label = 0)
    \item Personal Attack: ``Donald thinks belittling women makes him bigger. He goes after their dignity, their self-worth..." (Label = 1)
\end{itemize}

The second example was classified as a personal attack because it targets personal traits rather than political positions or policies (based on guideline 2).

\subsubsection{Exploratory Data Analysis}
The final dataset contained 2,239 sentences, of which 350 were labelled as personal attacks and 1,889 as non-attacks. There is a class imbalance because political debates focus more on policies and rebuttals rather on personal criticism. This imbalance highlighted the risk of models being biased toward the majority class. Attack rates vary from debate to debate because tone, topics and moderation style change. The highest rate of 27.2\% is in the 27 June 2024 debate which suggests a more confrontational exchange. The lowest rate of 12.6\% is in the 29 September 2020 debate indicating a more policy-focused discussion. These variations mean that a model trained on one debate may not perform equally well on another if the style shifts significantly. Table~\ref{tab:eda_by_debate} provides summary statistics in terms of \% of personal attacks across various debates. 

\begin{table}[]
  \centering
  \setlength{\arrayrulewidth}{0.6pt}
  \caption{Label counts and personal attack by debate.}
  \label{tab:eda_by_debate}
  \begin{tabular}{|l|c|c|c|}
    \hline
    \textbf{Debate} & \textbf{Non-attack} & \textbf{Attack} & \textbf{Attack \%} \\
    \hline
    09 Oct 2016   & 254 & 37  & 12.7 \\
    19 Oct 2016   & 266 & 68  & 20.4 \\
    29 Sep 2020 & 748 & 108 & 12.6 \\
    22 Oct 2020   & 301 & 50  & 14.2 \\
    27 Jun 2024      & 131 & 49  & 27.2 \\
    10 Sep 2024 & 189 & 38  & 16.7 \\
    \hline
  \end{tabular}
\end{table}

Year-wise totals also shift because candidate line-ups and main political issues change. In 2016 there are 105 attacks, in 2020 the number rises to 158 and in 2024 it drops to 87. This highlights how election-year dynamics can alter the overall tone of debates.

Sentence lengths vary widely because speakers alternate between short, pointed remarks and longer, detailed policy statements. The median is 57 characters (about 11 words). Shorter sentences often contain direct attacks while longer sentences usually explain policies. These observations guided preprocessing decisions, such as tokenisation and class weighting during training.

\subsubsection{Trump - Republican Candidate in 2016, 2020 and 2024}

Trump spoke most often about themes of \textit{economic focus}, \textit{repetition}, \textit{political attacks} and \textit{security concerns}. Monetary terms such as \textbf{``millions dollars''}, \textbf{``billions dollars''} and \textbf{``half million dollars''} appear often showing a focus on economic scale. Trigrams like \textbf{``make america great''} and \textbf{``going make america''} carry his campaign message and economic vision.

Repetition is another clear feature. Bigrams such as \textbf{``don know''}, \textbf{``ve seen''} and \textbf{``let just''} show an informal, assertive style; trigrams like \textbf{``let just tell''} and \textbf{``let just say''} reinforce a directive way of speaking.

Political references are frequent. Mentions of \textbf{``hillary clinton''}, \textbf{``33 000 mails''} and \textbf{``locker room talk''} point to controversial issues and attacks on opponents. Security and migration concerns also stand out with trigrams like \textbf{``people pouring country''} and \textbf{``people come country''} highlighting immigration-related narratives.

These patterns appear because Trump maintained a strong, threat-based narrative over three election cycles, shifting the perceived source of threat from \textbf{terror}, to \textbf{cities}, to \textbf{migrants}.

\subsubsection{Democratic Candidates in 2016, 2020 and 2024}

Democratic candidates spoke most often about \textit{healthcare}, \textit{national identity}, \textit{critiques of Trump} and \textit{social advocacy}. Healthcare is a central theme with frequent use of \textbf{``health care''} and \textbf{``affordable care''} and trigrams like \textbf{``affordable care act''} and \textbf{``pre existing conditions''} showing a consistent defense of Obama-era health reforms.

National identity and governance appear in phrases like \textbf{``united states''} and \textbf{``supreme court''} and trigrams such as \textbf{``president united states''} and \textbf{``united states senate''} which signals unity and respect for institutions.

Critiques of Trump are also common. The bigram \textbf{``donald trump''} and the trigram \textbf{``donald trump left''} are used to draw contrasts with his record and policies.

Social and economic advocacy comes through in phrases like \textbf{``middle class''}, \textbf{``middle class families''}, \textbf{``good paying jobs''} and \textbf{``pay fair share''} which points to a focus on economic fairness. Judicial and reproductive rights are highlighted through mentions of \textbf{``supreme court''} and related issues.

\begin{figure*}[]
    \centering
    \fbox{\includegraphics[scale=0.55]{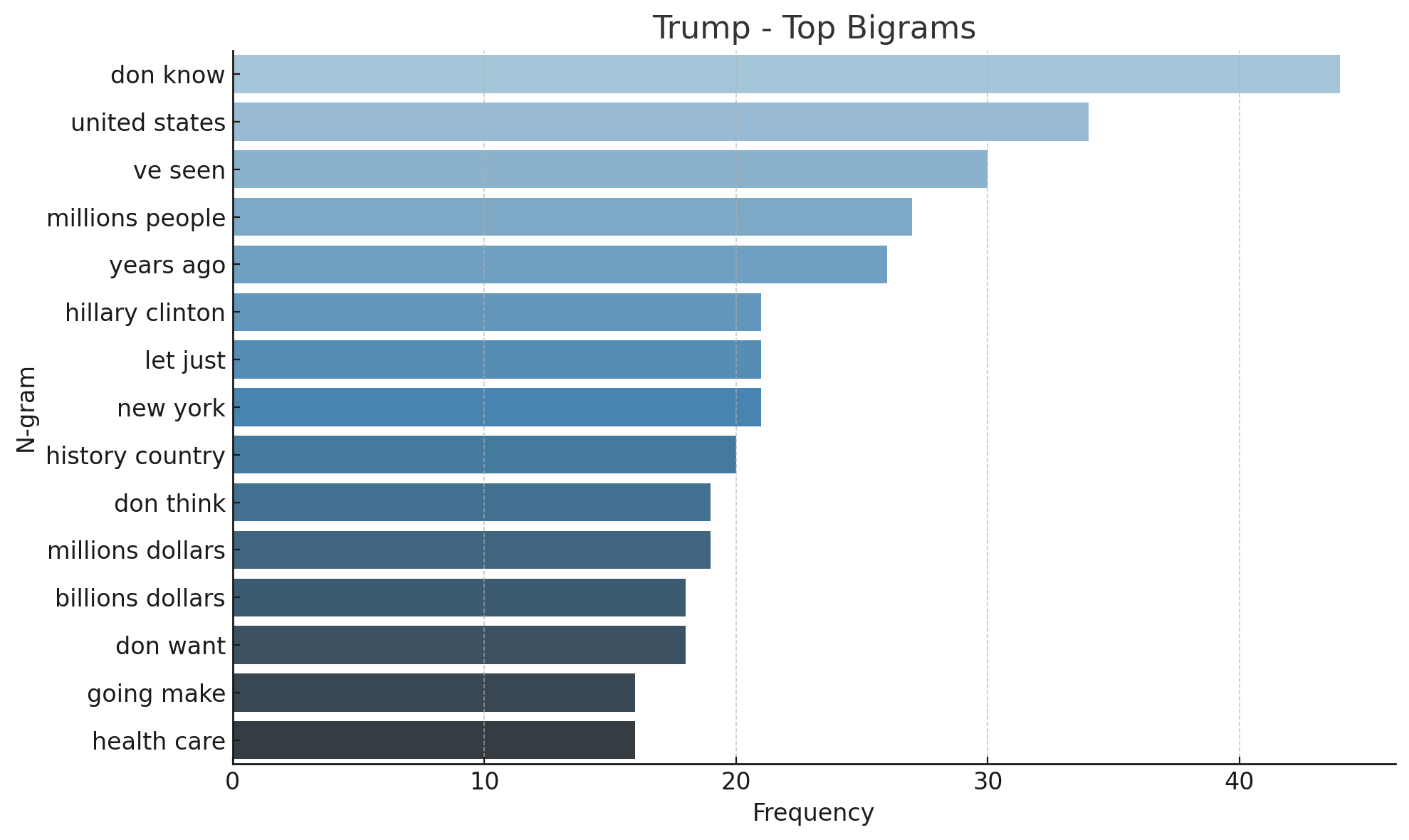}}
    \caption{Republicans – Top Bigrams across 2016, 2020 and 2024 debates.}
    \label{fig:republicans_bigrams}
\end{figure*}

\begin{figure*}[]
    \centering
    \fbox{\includegraphics[scale=0.55]{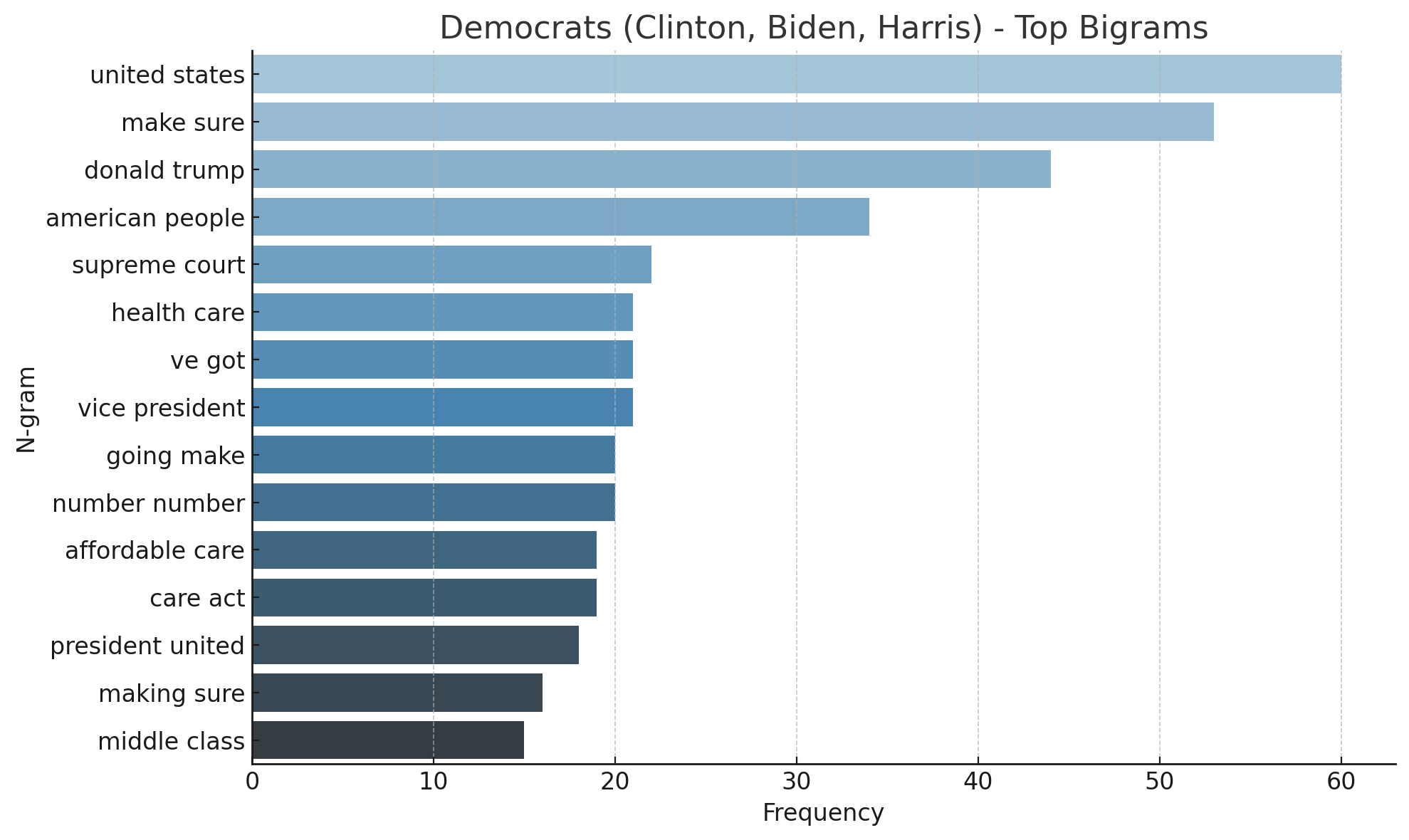}}
    \caption{Democrats– Top Bigrams across 2016, 2020 and 2024 debates.}
    \label{fig:democrats_bigrams}
\end{figure*}

\begin{figure*}[]
    \centering
    \fbox{\includegraphics[scale=0.55]{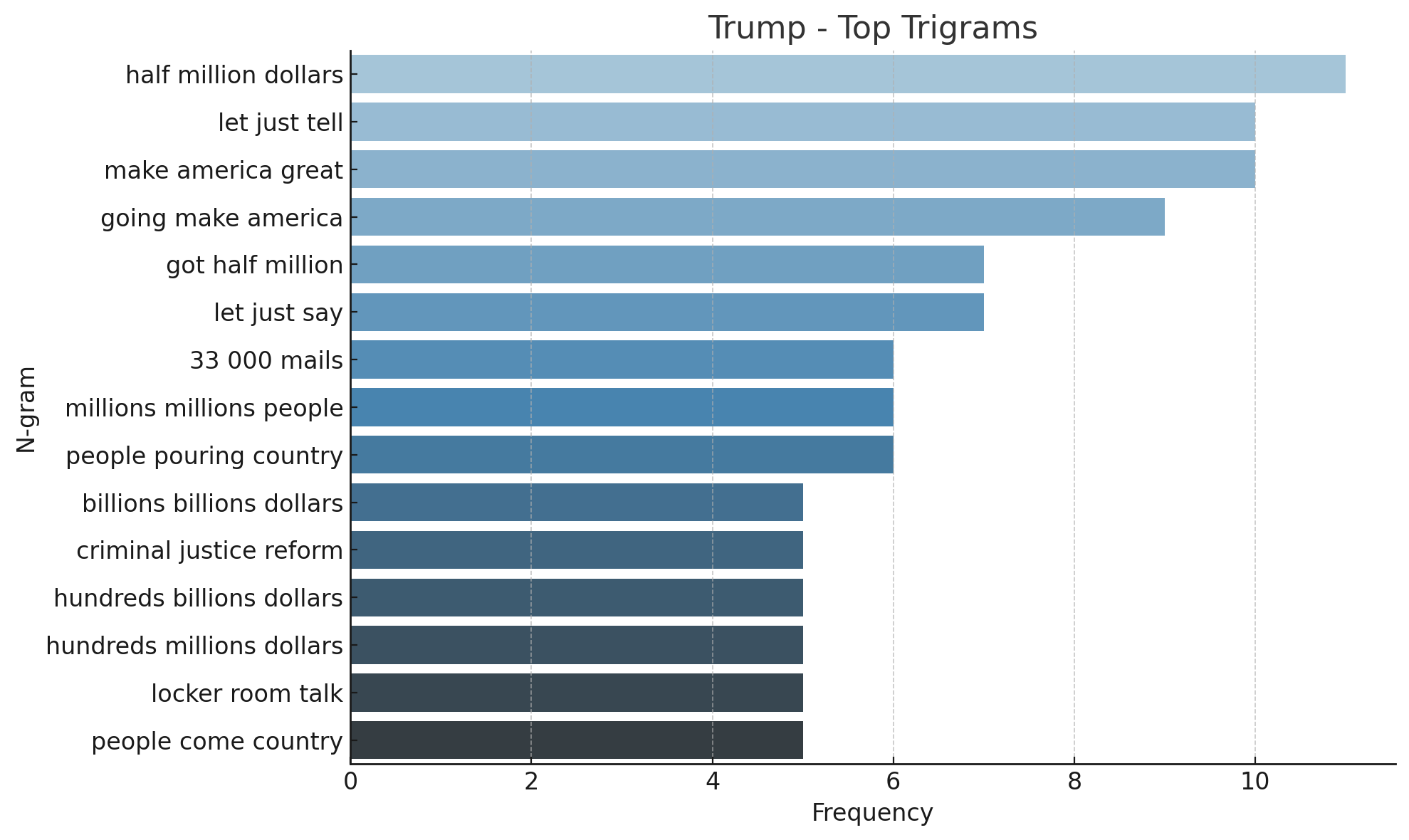}}
    \caption{Republicans – Top Trigrams across 2016, 2020 and 2024 debates.}
    \label{fig:republicans_trigrams}
\end{figure*}

\begin{figure*}[]
    \centering
    \fbox{\includegraphics[scale=0.55]{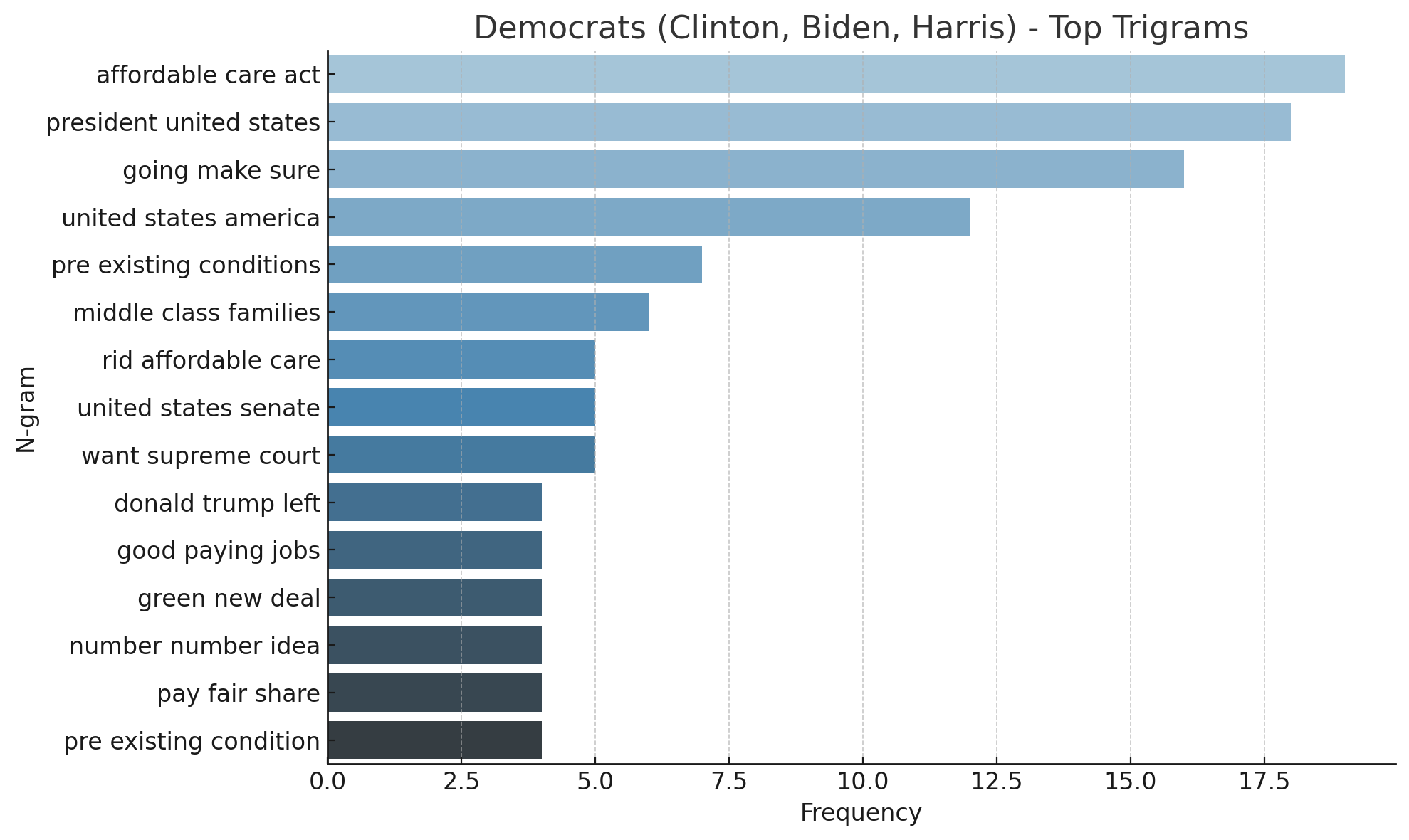}}
    \caption{Democrats – Top Trigrams across 2016, 2020 and 2024 debates.}
    \label{fig:dem_bigrams}
\end{figure*}

These analyses provided qualitative insight into rhetorical strategies and informed model training by highlighting linguistic differences across debates and years.

\section{Experiments}~\label{sec:experiments}

\subsection{Baseline model}
To establish a baseline, HateBERT\cite{HateBERT} was selected as the initial model for evaluation. HateBERT is a finetuned version of BERT that was trained on offensive and abusive language from Reddit communities, making it a strong candidate for detecting hostile language in text. The aim was to test whether such a model could directly be applied to the more formal language of US presidential debates without additional task-specific fine-tuning. The HateBERT model performed poorly, giving an accuracy of 21.13\%, precision of 15.90\%, recall of 94.29\%, and F1-score of 27.21\%. These results show that while the HateBERT model captured most of the actual personal attacks (high recall), it also incorrectly classified a large number of non-attacks as attacks (low precision), resulting in poor overall accuracy.

The evaluation showed that HateBERT struggled to detect subtle or indirect personal attacks that are common in presidential debates. Its Reddit-based training helped to spot explicit abuse but it was less effective at recognising sarcasm, veiled insults and implied criticism. Because the training data used informal and direct language while the target domain relied on formal political speech, many statements were misclassified. These results show that although HateBERT works well for informal online hate speech, it is not suited to the nuanced language of political debates. This highlighted the need for domain-specific fine-tuning using the annotated debate dataset to better capture patterns in political discourse.

\subsection{Fine-tuning BERT model for personal attack prediction}
After finding that HateBERT struggled with the formal and nuanced language of political debates, we shifted our focus to fine-tune general-purpose BERT models on the annotated dataset. The aim was to adapt the model to the context-driven style of US presidential debates and improve its ability to detect subtle personal attacks. We setup our experiments based on two settings, allowing us to measure both how well the model learned from mixed debates and how well it adapted to a completely new one. Table~\ref{tab:bert_configuration} provides various hyper-parameters for fine-tuning BERT model. 

\begin{table*}[]
    \centering
    \footnotesize
    \caption{Fine-tuned BERT Configuration}
    \label{tab:bert_configuration}
    \begin{tabular}{p{2.4cm}p{4cm}p{4cm}p{4.5cm}}
    \hline
    \textbf{Category} & \textbf{Parameter} & \textbf{Value} & \textbf{Purpose/Impact} \\ \hline 
    Training split & Train val test split ratio & 80\%,10\%,10\% (only setup 1) & Standard split ensuring model validation and testing \\
    Training Split & Seeded and Stratified Split & Enabled & Preserves class distribution across all splits \\
    Model & Base Model & \texttt{bert-base-uncased} & Pre-trained English BERT model \\
    Model & Dropout (hidden \& attention) & 0.3 & Helps prevent overfitting \\
    Model & Num Labels & 2 & Binary classification \\
    Loss Function & \texttt{CrossEntropyLoss (weight=class\_weights)} & \texttt{Inverse-frequency class weights} & Balances class imbalance in the loss function \\
    Class Weights & Inverse Frequency & \texttt{[1/class0\_freq, 1/class1\_freq]} & Penalises the minority class to reduce bias \\
    Optimizer & AdamW & Yes & Commonly used with Transformers \\
    Optimizer & Weight Decay & 0.05 & Regularisation to prevent overfitting \\
    Optimizer & Learning Rate & $1\times10^{-5}$ & Stable fine-tuning \\
    Scheduler & Type & Linear scheduler & Gradually decreases learning rate \\
    Batch Sizes & Train/Eval & 16 & Balance of memory use and stability \\
    Batch Sizes & Test & 8 & Efficient test inference \\
    Training Epochs & Max Epochs & 10 & Max training duration \\
    Training Epochs & Early Stopping Patience & \texttt{2 epochs without F1 improvement} & Prevents overfitting \\

    \hline
    \end{tabular}
\end{table*}

\subsubsection{Setup 1: Combined Debate Training (80/10/10 split)}

In this setup, all annotated debate transcripts were combined and split into training, validation and test sets in an 80/10/10 ratio. The split was both seeded and stratified, ensuring the same distribution of personal attack and non-attack instances in each set while keeping the results reproducible. This allowed the model to learn from a diverse mix of debates while still having balanced class representation.

\begin{figure}[]
    \centering
    \includegraphics[scale=0.45]{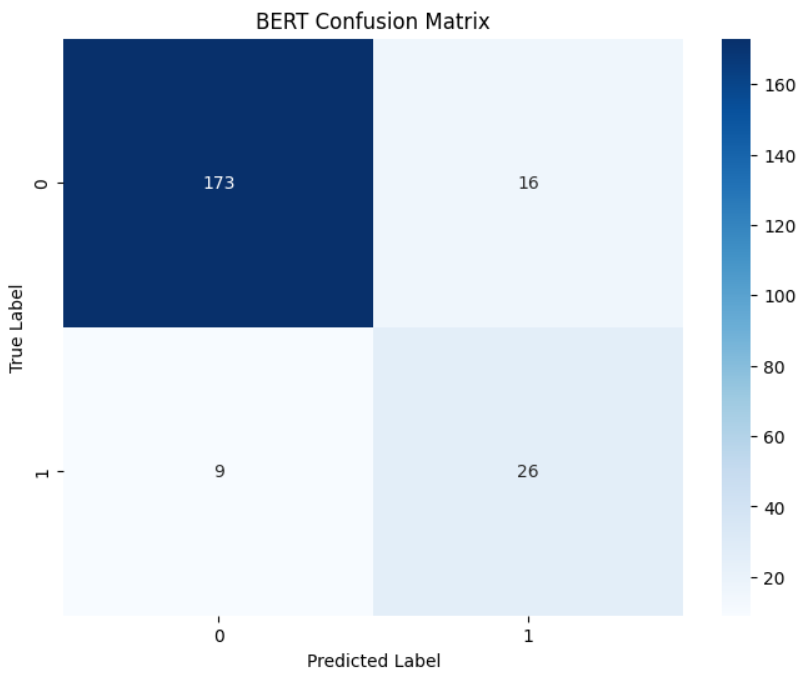}
    \caption{Confusion Matrix for BERT Model (80/10/10) on the test set.}
    \label{fig:model1_confusion}
\end{figure}

On the test set, the model achieved an accuracy of 88.84\% and an F1-Score of 89.24\%. Precision of 89.88\%, Recall of 88.84\% and the AUC of 0.93 confirmed strong separability between personal attacks and non-attacks. Out of all personal attack cases, 26 were correctly identified while only nine were missed.  
Figure~\ref{fig:model1_confusion} shows that most predictions lie along the diagonal with very few false positives and false negatives.

\subsubsection{Setup 2: Leave-One-Debate-Out Evaluation}
\begin{figure}[]
    \centering
    \includegraphics[scale=0.45]{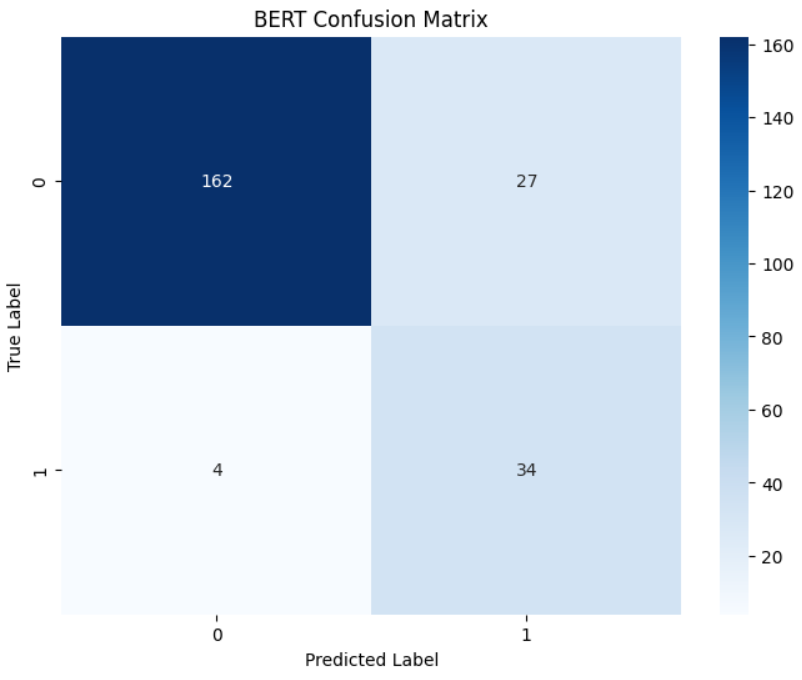}
    \caption{Confusion Matrix for BERT Model (Test: 2024) }
    \label{fig:model2_confusion}
\end{figure}

In this setup, debates from 2016 and 2020 were used for training and validation, while one full debate transcript from 2024 was held out as the test set. This tested the model’s ability to handle an entirely unseen debate, similar to a real-world case where the language, tone or topics might differ from past data. Figure~\ref{fig:model2_confusion} shows the confusion matrix for the BERT model predictions on the 2024 debate. The model achieved an accuracy of 86.34\% with 34 personal attacks correctly identified and only 4 missed. 
The Precision of 90.58\% shows the model rarely labels neutral statements as attacks, while the Recall of 86.34\% indicates strong coverage even on unfamiliar data. 
Although 27 neutral statements were misclassified, the AUC score of 92.33\% confirms that the model maintained good class separation when handling new election contexts. 

\subsubsection{Comparison}
The results of comparing BERT models in the above two settings is given in Table~\ref{tab:bert_results}.
The BERT Model (80/10/10) outperformed BERT Model (Test: 2024) in most metrics with higher Accuracy, Recall, F1 Score and AUC. This indicated stronger overall consistency and balance between Precision and Recall. However, the BERT Model (Test: 2024) achieved a slightly higher Precision, suggesting it was more selective in identifying personal attacks and predicted fewer false positives.

\begin{table}[]
\centering
\caption{Performance comparison between BERT Model (80/10/10) and BERT Model (Test: 2024). Higher values are shown in bold.}
\label{tab:bert_results}
\begin{tabular}{p{2cm}p{2.2cm}p{2.2cm}}
\hline
\textbf{Metric} & \textbf{BERT Model (80/10/10)} & \textbf{BERT Model (Test: 2024)} \\ \hline
Accuracy & \textbf{0.8884} & 0.8634 \\ 
Precision & 0.8988 & \textbf{0.9058} \\ 
Recall & \textbf{0.8884} & 0.8634 \\ 
F1 Score & \textbf{0.8924} & 0.8749 \\ 
AUC & \textbf{0.9262} & 0.9233 \\ \hline
\end{tabular}
\end{table}

\subsection{LLM-based prediction and fine-tuning}
After we fine-tuned BERT model, we focused on answering the question how large language models (LLMs) could be used both in their base form and after fine-tuning to detect personal attacks in US presidential debate transcripts. The prediction phase focused on evaluating multiple leading LLMs on a shared test set while the fine-tuning phase examined whether adapting a single model to this specific task could improve its Accuracy and robustness. Together, these approaches provided insight into how well general-purpose models can handle the nuanced language of political debates and the extent to which targeted training can enhance their performance.

We evaluated the ability of leading Large Language Models (LLMs), namely, \emph{ChatGPT-4o}, \emph{DeepSeek-V3}, \emph{Claude Sonnet 4}, \emph{Gemini 2.5 Pro}, and \emph{Grok 3}, to detect personal attacks in political debate text, using a consistent and controlled test set of 224 sentences (exactly as the test set for BERT models).
Each model received the same instruction, which required it to: (1) assign a \textbf{binary label}: 1 for personal attack, 0 for non-attack; and (2) provide a \textbf{brief justification} for each classification. This approach allowed for both quantitative evaluation (standard classification metrics) and qualitative analysis (reviewing justification text). For all the models, the following \emph{prompt} was used:

\begin{tcolorbox}[colback=gray!5, colframe=black, sharp corners, boxrule=0.5pt, left=5pt, right=5pt, top=5pt, bottom=5pt]
You will be given text in the following prompt containing rows from US Presidential debate transcripts. 
Your task is to read each row and classify the content of the sentence using binary annotation based on whether it contains a personal attack.

A personal attack refers to a remark that targets a person’s character, intelligence, appearance, morals, background or any trait unrelated to the topic being debated. 
It includes insults, name-calling, mockery or any language meant to undermine the individual personally.

In contrast, non-personal attacks focus on ideas, policies or arguments, even if critical or strongly worded.

For each row (sentence), provide:
\begin{enumerate}
    \item A binary label — 1 if the sentence contains a personal attack, 0 otherwise.
    \item A short justification explaining why you gave that label.
\end{enumerate}

Add the binary label in a new column named after your model (e.g., ChatGPT) and the justification in another column named \texttt{Justification}.
\end{tcolorbox}

\subsubsection{ChatGPT-4o}
ChatGPT-4o achieved an overall accuracy of 84.82\% and showed moderate ability to detect personal attacks in political debate text. However, its precision of 51.85\% and recall of 40.00\% reveal a mismatch. While it was fairly good at avoiding false positives, it often failed to catch genuine attacks. This under-detection is reflected in its F1-score of 45.16\%.

\begin{figure}[]
    \centering
    \includegraphics[scale=0.55]{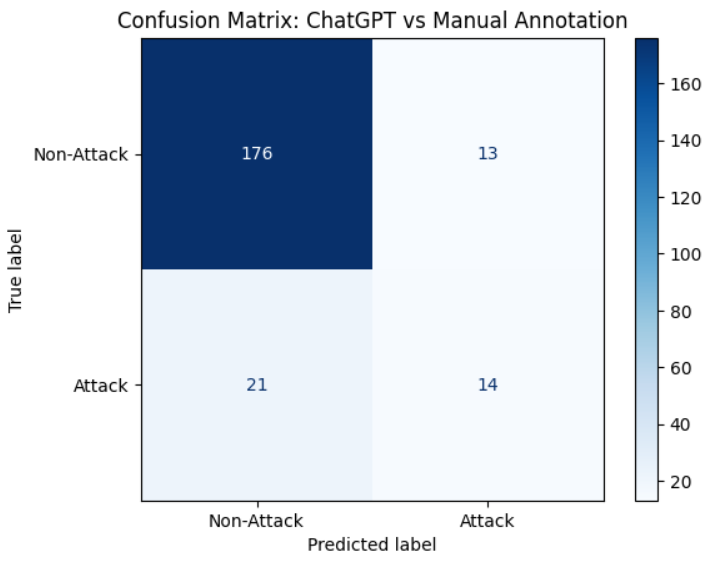}
    \caption{Confusion Matrix – ChatGPT-4o (Test Set Predictions)}
    \label{fig:cm_chatgpt4o}
\end{figure}

A closer review of its justifications showed they were often \textit{template-based} and relied heavily on keywords like ``you're,'' ``disgrace,'' or ``he is'' instead of considering the full sentence context. This keyword-focused approach meant it frequently missed subtle or implied attacks which reduced both recall and the variety of its explanations. Figure~\ref{fig:cm_chatgpt4o} shows the confusion matrix, highlighting the clear gap between true positives and false negatives.

\subsubsection{Claude Sonnet 4}

Claude Sonnet 4 achieved an accuracy of 91.96\% with a precision of 71.79\%, recall of 80.00\% and an F1-score of 75.68\% making it the most balanced model so far. Its predictions were supported by explanations that were more \textit{context-aware} and \textit{detailed} compared to ChatGPT. For example, it flagged lines implying corruption, questioning competence or hinting at questionable dealings, even when such attacks were not explicitly worded.   

While not perfect, Claude’s annotations reflected more \textit{human-like reasoning}, considering tone, implication and the speaker’s role (e.g., moderator vs. candidate). This allowed it to balance Precision and Recall more effectively. This made Claude a stronger performer as compared to ChatGPT.

\begin{figure}[]
    \centering
    \includegraphics[scale=0.55]{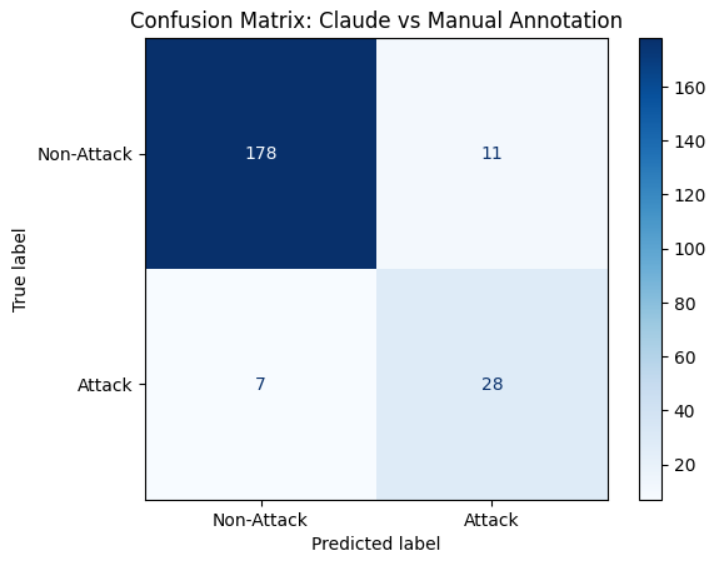}
    \caption{Confusion Matrix – Claude Sonnet 4 (Test Set Predictions)}
\end{figure}

\subsubsection{DeepSeek-V3}

DeepSeek-V3 delivered the highest accuracy so far at 92.86\% with a precision of 71.11\%, recall of 91.43\% and an F1-score of 80.00\%. Its predictions were backed by clear and \textit{context-aware} reasoning. For example, it flagged phrases implying corruption or questioning competence, even when no direct insult was present.  

This strong contextual understanding enabled DeepSeek-V3 to detect a wide range of personal attacks from overt remarks to subtle implications. Its Recall (91.43\%) was notably higher than Claude’s 80.00\%, showing greater sensitivity to attacks. However, its slightly lower Precision suggests a marginally higher false positive rate. Overall, DeepSeek-V3 balanced strong detection with contextual reasoning. This made DeepSeek one of the most effective LLMs evaluated so far.

\begin{figure}[]
    \centering
    \includegraphics[scale=0.55]{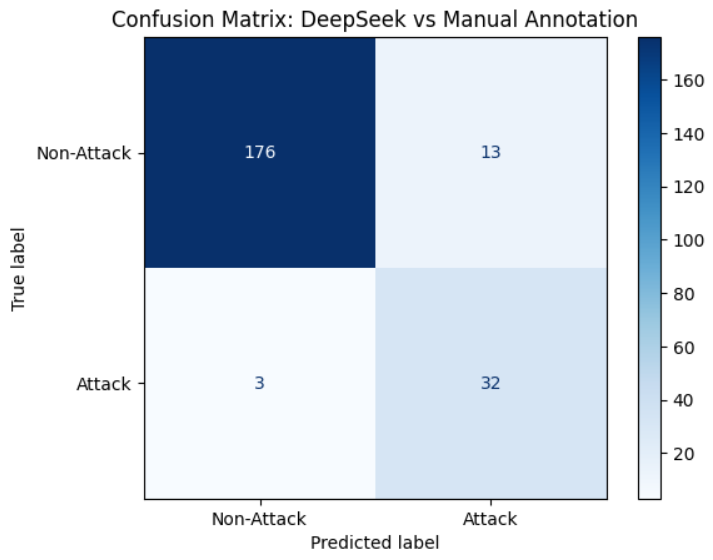}
    \caption{Confusion Matrix – DeepSeek-V3 (Test Set Predictions)}
\end{figure}

\subsubsection{Gemini 2.5 Pro}
Gemini 2.5 Pro achieved an accuracy of 91.52\% with a precision of 72.22\%, recall of 74.29\% and an F1-score of 73.24\%. Its justifications were \textit{detailed} and \textit{focused}. For example, it identified moral or integrity-based attacks while dismissing non-attacks.

Gemini also showed strength in distinguishing \textit{policy critique} from \textit{personal attack}. Compared to DeepSeek-V3, Gemini had slightly lower recall (74.29\% vs. 91.43\%), meaning it missed more actual attacks. However, it maintained \textbf{higher Precision} showing better control over false positives.

\begin{figure}[]
    \centering
    \includegraphics[scale=0.55]{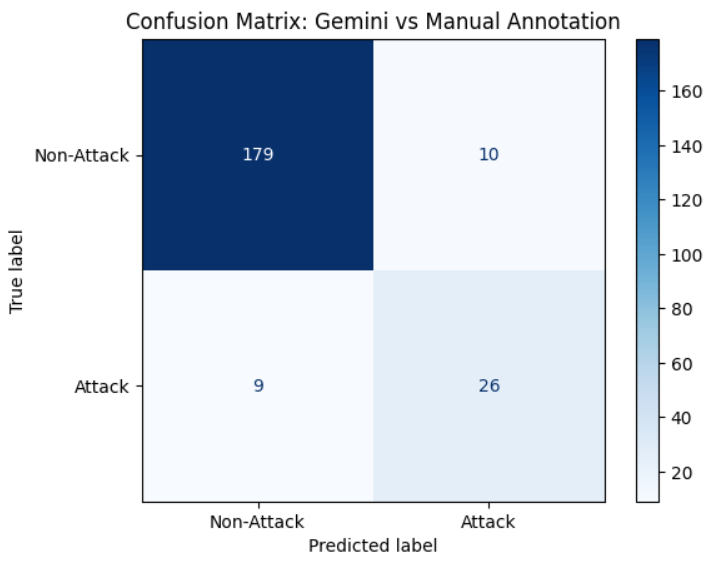}
    \caption{Confusion Matrix – Gemini 2.5 Pro (Test Set Predictions)}
\end{figure}

\subsubsection{Grok 3}

Grok 3 achieved an accuracy of 90.62\% with a precision of 64.00\%, recall of 91.43\% and an F1 Score of 75.29\%. Its justifications were \textit{contextually aware}, balancing speaker intent, moderation and implied meaning. For example, it flagged competence-based attacks (It marked those sentences as personal attacks which implied a lack of competence) while dismissing neutral remarks.  

Compared to Gemini, Grok achieved much higher Recall (91.43\% vs. 74.29\%), capturing more true personal attacks. Overall, Grok leaned toward \textit{over-detection}, making it more suitable for scenarios where missing potential attacks is riskier than allowing a few extra false positives.

\begin{figure}[]
    \centering
    \includegraphics[scale=0.55]{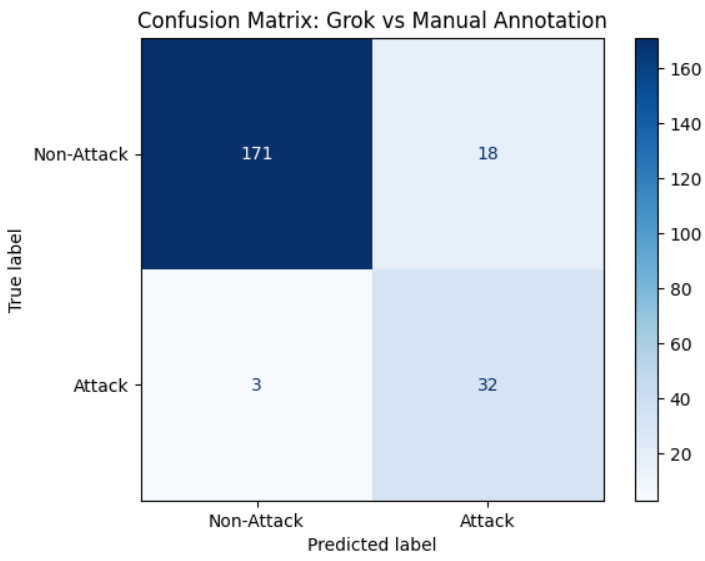}
    \caption{Confusion Matrix – Grok 3 (Test Set Predictions)}
\end{figure}

\subsubsection{Comparison of various LLMs-based predictions}
Table~\ref{tab:llm-quant} presents key metrics for each model. DeepSeek-V3 achieved the highest Accuracy and Recall. Gemini had the highest Precision and Claude provided a good balance across all metrics.

\begin{table*}[]
\centering
\caption{Quantitative comparison of LLMs on personal attack detection.}
\label{tab:llm-quant}
\begin{tabular}{|l|c|c|c|c|}
\hline
\textbf{Model} & \textbf{Accuracy} & \textbf{Precision} & \textbf{Recall} & \textbf{F1-score} \\
\hline
ChatGPT-4o & 0.8482 & 0.5185 & 0.4000 & 0.4516 \\
Claude Sonnet 4 & 0.9196 & 0.7179 & 0.8000 & 0.7568 \\
Gemini 2.5 Pro & 0.9152 & \textbf{0.7222} & 0.7429 & 0.7324 \\
DeepSeek-V3 & \textbf{0.9286} & 0.7111 & \textbf{0.9143} & \textbf{0.8000} \\
Grok 3 & 0.9062 & 0.6400 & \textbf{0.9143} & 0.7529 \\
\hline
\end{tabular}
\end{table*}

Table summarises the reasoning styles of each model when detecting personal attacks. DeepSeek-V3 and Grok focus on high Recall. Gemini 2.5 Pro applies a Precision-oriented approach. Claude Sonnet 4 offers balanced and context-aware reasoning. ChatGPT-4o provides consistent and concise outputs.

\begin{table*}[]
\centering
\caption{Qualitative comparison of LLM reasoning styles for personal attack detection.}
\label{tab:llm-reasoning}
\begin{tabular}{ll}
\hline
\textbf{Model} & \textbf{Reasoning Style} \\
\hline
ChatGPT-4o & Keyword-driven, concise justifications \\
Claude Sonnet 4 & Context-aware, explains tone and intent \\
Gemini 2.5 Pro & Focused, uses debate roles and context \\
DeepSeek-V3 & Clear, context-rich explanations \\
Grok 3 & Sensitive to implied meaning \\
\hline
\end{tabular}
\end{table*}

\subsection{Fine-tuning LLaMa}
The off-the-shelf LLMs predictions could not consistently detect both explicit and implied personal attacks in political debates. Some LLMs missed subtle context-driven cases while others misclassified neutral statements as attacks. These patterns point to a mismatch between the open-domain training of these models and the specialised linguistic style of political debates. To investigate whether domain adaptation could address these limitations, we fine-tuned \emph{LLaMA-3.2-3B-Instruct} model on the annotated debate dataset.

\begin{figure*}[]
\centering
\includegraphics[width=0.95\textwidth]{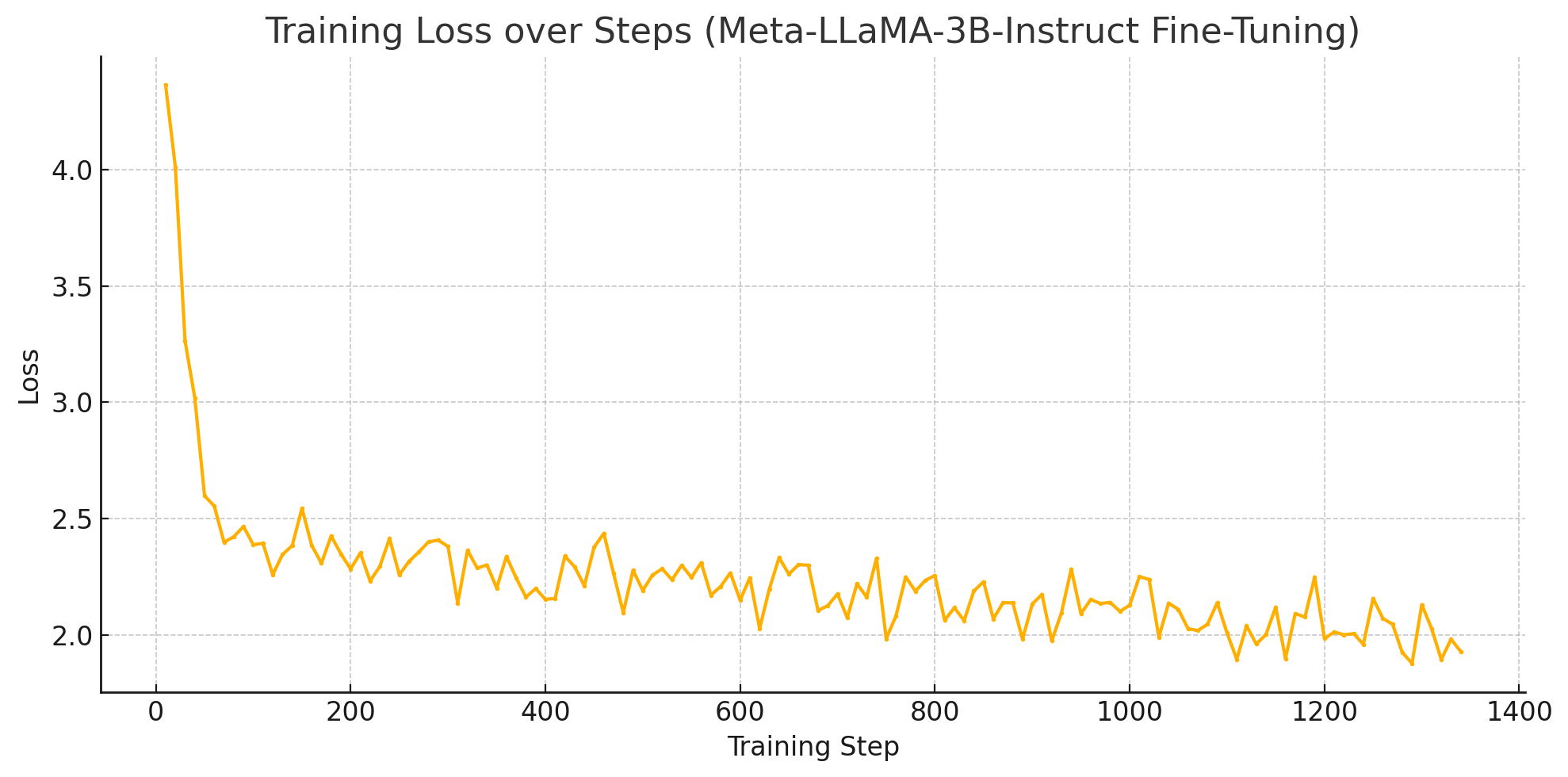}
\caption{Training loss over steps for Meta-LLaMA-3B-Instruct fine-tuning \\
(until step 1340).}
\label{fig:llm_training_loss}
\end{figure*}

The \emph{Meta-LLaMA-3B-Instruct} model was fine-tuned using the same train and validation set as the seeded BERT model to maintain a consistent evaluation setup. Training used \emph{LoRA (Low-Rank Adaptation)} with 4-bit quantisation for memory-efficient adaptation without updating all parameters. The process ran on a NVIDIA A100 GPU for 3 epochs and 1344 steps reaching a final training loss of 2.23. The training loss dropped from 4.36 to 1.93, showing steady convergence and stable learning. Figure \ref{fig:llm_training_loss} illustrates the loss curve with a sharp initial decline followed by gradual stabilisation.

We compared the base \texttt{Meta-LLaMA-3B-Instruct} model with its fine-tuned version.
The \emph{base model} reached 65.36\% accuracy. It performed well on the “Not Attack” class but had low precision (0.21) for “Attack” and often misclassified non-attacks as attacks. This resulted in 46 false positives and 16 false negatives. The \emph{fine-tuned model} improved precision to 0.26 and reduced false positives to 31. It balanced “Attack” predictions better but recall dropped to 0.39 and skipped predictions increased from 45 to 86. Accuracy remained similar at 65.22\%.

\section{Discussion}
This study demonstrated that task-specific fine-tuning of transformer-based and large language models improves the detection of personal attacks in presidential debates. However, there are several directions for future work:

\begin{itemize}
    \item Larger and Diverse Datasets: Expanding the dataset with more annotated debates and political speeches would help improve generalisation. Including data from congressional debates, campaign rallies, and interviews could provide richer context.
    \item Multilingual Capabilities: Extending the framework to non-English debates would allow comparative analysis across different political systems and cultural settings, highlighting how personal attacks vary globally.
    \item Explainability and Interpretability: Future work can explore explainable AI approaches to show why a statement is classified as a personal attack. This would increase trust and usability for journalists and analysts.
    \item Larger and Advanced Models: Fine-tuning larger LLMs such as GPT-4-class or Gemini Ultra could further improve detection accuracy and reasoning depth, provided computational resources allow it.
    \item Integration with Media Tools: Developing applications that integrate these models into real-time debate analysis platforms could provide immediate insights for the public and media outlets.
\end{itemize}

\section{Conclusion}

This paper explored how transformer-based models and LLMs can be applied to detect personal attacks in U.S. presidential debates. A manually annotated dataset of transcripts from the 2016, 2020, and 2024 debates was created and analysed. Initial tests with HateBERT highlighted the limits of applying existing hate speech models to formal political contexts. Fine-tuned BERT models achieved strong results, showing that task-specific training improves both accuracy and generalisation.  LLMs such as ChatGPT, Claude, Gemini, Grok, and DeepSeek were also evaluated. Each showed complementary strengths, with trade-offs between precision and recall. Finally, fine-tuning the Meta-LLaMA-3B-Instruct model with LoRA confirmed that domain-specific adaptation further improves performance.  Overall, the findings demonstrate that automated detection of personal attacks is feasible and effective when models are adapted to political discourse. This contributes to more transparent political analysis and provides a foundation for future research in multilingual, explainable, and real-time applications of AI in political communication.

\section*{Code and Data}

GitHub repository for code and data \url{https://github.com/rubangoyal03/Analysing-Personal-Attacks-in-U.S.-Presidential-Debates}
\section{Acknowledgements}

This research was supported by Katana, the high performance computing facility at the University of New South Wales. The authors also  acknowledge the financial support provided by the School of Computer Science and Engineering for API and cloud services used for this study.

\section{Bibliographical References}\label{sec:reference}

\bibliographystyle{lrec2026-natbib}
\bibliography{lrec2026-references}

@article{khurana2023natural,
  title={{Natural Language Processing: State of the Art, Current Trends and Challenges}},
  author={Khurana, Dhruv and Koli, Ashlesha and Khatter, Kiran and Singh, Sukhdev},
  journal={Multimedia Tools and Applications},
  volume={82},
  number={3},
  pages={3713--3744},
  year={2023},
  publisher={Springer},
  doi={10.1007/s11042-022-13428-4},
  url={https://doi.org/10.1007/s11042-022-13428-4}
}

@article{DBLP:journals,
  author       = {Jacob Devlin and
                  Ming{-}Wei Chang and
                  Kenton Lee and
                  Kristina Toutanova},
  title        = {{BERT: Pre-training of Deep Bidirectional Transformers for Language
                  Understanding}},
  journal      = {CoRR},
  volume       = {abs/1810.04805},
  year         = {2018},
  url          = {http://arxiv.org/abs/1810.04805},
  doi          = {10.48550/arXiv.1810.04805}
}

@article{ray2023chatgpt,
  title={{ChatGPT: A Comprehensive Review on Background, Applications, Key Challenges, Bias, Ethics, Limitations and Future Scope}},
  author={Ray, P. P.},
  journal={Internet of Things and Cyber-Physical Systems},
  volume={3},
  pages={121--154},
  year={2023},
  doi={10.1016/j.iotcps.2023.04.003},
  url={https://doi.org/10.1016/j.iotcps.2023.04.003}
}

@article{imran2024google,
  title={{Google Gemini as a Next Generation AI Educational Tool: A Review of Emerging Educational Technology}},
  author={Imran, Muhammad and Almusharraf, Norah},
  journal={Smart Learning Environments},
  volume={11},
  number={22},
  year={2024},
  doi={10.1186/s40561-024-00310-z},
  url={https://doi.org/10.1186/s40561-024-00310-z}
}

@inproceedings{cavnar1994n,
  title={{N-gram Based Text Categorization}},
  author={Cavnar, William B and Trenkle, John M and others},
  booktitle={Proceedings of SDAIR-94, 3rd annual symposium on document analysis and information retrieval},
  volume={161175},
  pages={14},
  year={1994}, 
  url= {https://www.researchgate.net/publication/2375544_N-Gram-Based_Text_Categorization}
}

@article{HateBERT,
  author       = {Tommaso Caselli and
                  Valerio Basile and
                  Jelena Mitrovic and
                  Michael Granitzer},
  title        = {{HateBERT: Retraining BERT for Abusive Language Detection in English}},
  journal      = {CoRR},
  volume       = {abs/2010.12472},
  year         = {2020},
  url          = {https://arxiv.org/abs/2010.12472},
  doi          = {10.48550/arXiv.2010.12472}
}

@article{BidenvsTrump,
  author={Chandra, Rohitash and Saini, Ritij},
  journal={IEEE Access}, 
  title={{Biden vs Trump: Modeling US General Elections Using BERT Language Model}}, 
  year={2021},
  volume={9},
  pages={128494--128505},
  doi={10.1109/ACCESS.2021.3111035},
  url={https://doi.org/10.1109/ACCESS.2021.3111035}
}

@mastersthesis{fluit2020polarization,
  title={{Polarization and Personal Attacks in American Presidential Debates: A study of the use of ad hominem arguments in the American Presidential Debates leading up to the presidencies of Barack Obama in 2008 and Donald Trump in 2016}},
  author={Fluit, Laurens},
  school={Linguistics – Language \& Communication},
  year={2020},
  month={January},
  url={https://studenttheses.universiteitleiden.nl/access/item%3A2630222/view},
  note={{Supervisor: Roosmaryn Pilgram; Second reader: Ton van Haaften}}
}

@inproceedings{gladwin2022toxicbert,
  title={{Toxic Comment Identification and Classification using BERT and SVM}},
  author={Gladwin, Ivander and Renjiro, Evan Vitto and Valerian, Bryan and Edbert, Ivan Sebastian and Suhartono, Derwin},
  booktitle={2022 8th International Conference on Science and Technology (ICST)},
  year={2022},
  pages={1--6},
  publisher={IEEE},
  doi={10.1109/ICST56971.2022.10136295},
  url={https://doi.org/10.1109/ICST56971.2022.10136295}
}

@ARTICLE{Singh:2025:HP_BERT,
  author={Singh, Ashutosh and Chandra, Rohitash},
  journal={IEEE Access}, 
  title={HP-BERT: A framework for longitudinal study of Hinduphobia on social media via language models}, 
  year={2025},
  volume={},
  number={},
  pages={1-1},
  doi={10.1109/ACCESS.2025.3617514}
}

@article{touvron2023llama,
      title={LLaMA: Open and Efficient Foundation Language Models}, 
      author={Hugo Touvron and Thibaut Lavril and Gautier Izacard and Xavier Martinet and Marie-Anne Lachaux and Timothée Lacroix and Baptiste Rozière and Naman Goyal and Eric Hambro and Faisal Azhar and Aurelien Rodriguez and Armand Joulin and Edouard Grave and Guillaume Lample},
      year={2023},
      journal={arXiv preprint arXiv:2302.13971},
}

@misc{LoRA,
      title={LoRA: Low-Rank Adaptation of Large Language Models}, 
      author={Edward J. Hu and Yelong Shen and Phillip Wallis and Zeyuan Allen-Zhu and Yuanzhi Li and Shean Wang and Lu Wang and Weizhu Chen},
      year={2021},
      eprint={2106.09685},
      archivePrefix={arXiv},
      primaryClass={cs.CL},
      url={https://arxiv.org/abs/2106.09685}, 
}

@inproceedings{devlin-etal-2019-bert,
    title = "{BERT}: Pre-training of Deep Bidirectional Transformers for Language Understanding",
    author = "Devlin, Jacob  and
      Chang, Ming-Wei  and
      Lee, Kenton  and
      Toutanova, Kristina",
    editor = "Burstein, Jill  and
      Doran, Christy  and
      Solorio, Thamar",
    booktitle = "Proceedings of the 2019 Conference of the North {A}merican Chapter of the Association for Computational Linguistics: Human Language Technologies, Volume 1 (Long and Short Papers)",
    month = jun,
    year = "2019",
    address = "Minneapolis, Minnesota",
    publisher = "Association for Computational Linguistics",
    url = "https://aclanthology.org/N19-1423/",
    doi = "10.18653/v1/N19-1423",
    pages = "4171--4186",
}

@article{openai2023gpt4,
  title={GPT-4 Technical Report},
  author={OpenAI},
  journal={arXiv preprint arXiv:2303.08774},
  year={2023}
}

@inproceedings{Brown2020gpt3,
 author = {Brown, Tom and Mann, Benjamin and Ryder, Nick and Subbiah, Melanie and Kaplan, Jared D and Dhariwal, Prafulla and Neelakantan, Arvind and Shyam, Pranav and Sastry, Girish and Askell, Amanda and Agarwal, Sandhini and Herbert-Voss, Ariel and Krueger, Gretchen and Henighan, Tom and Child, Rewon and Ramesh, Aditya and Ziegler, Daniel and Wu, Jeffrey and Winter, Clemens and Hesse, Chris and Chen, Mark and Sigler, Eric and Litwin, Mateusz and Gray, Scott and Chess, Benjamin and Clark, Jack and Berner, Christopher and McCandlish, Sam and Radford, Alec and Sutskever, Ilya and Amodei, Dario},
 booktitle = {Advances in Neural Information Processing Systems},
 editor = {H. Larochelle and M. Ranzato and R. Hadsell and M.F. Balcan and H. Lin},
 pages = {1877--1901},
 publisher = {Curran Associates, Inc.},
 title = {Language Models are Few-Shot Learners},
 url = {https://proceedings.neurips.cc/paper_files/paper/2020/file/1457c0d6bfcb4967418bfb8ac142f64a-Paper.pdf},
 volume = {33},
 year = {2020}
}

@misc{claude3,
    title={Introducing the next generation of Claude},
    author={Anthropic},
    year={2024},
    url={https://www.anthropic.com/news/claude-3-family}
}

@article{geminiteam2023gemini,
      title={Gemini: A Family of Highly Capable Multimodal Models}, 
      author={{Gemini Team Google}},
      year={2023},
      journal={arXiv preprint arXiv:2312.11805},
}

@misc{deepseekai2024,
      title={DeepSeek LLM: Scaling Open-Source Language Models with Longtermism}, 
      author={DeepSeek-AI and : and Xiao Bi and Deli Chen and Guanting Chen and Shanhuang Chen and Damai Dai and Chengqi Deng and Honghui Ding and Kai Dong and Qiushi Du and Zhe Fu and Huazuo Gao and Kaige Gao and Wenjun Gao and Ruiqi Ge and Kang Guan and Daya Guo and Jianzhong Guo and Guangbo Hao and Zhewen Hao and Ying He and Wenjie Hu and Panpan Huang and Erhang Li and Guowei Li and Jiashi Li and Yao Li and Y. K. Li and Wenfeng Liang and Fangyun Lin and A. X. Liu and Bo Liu and Wen Liu and Xiaodong Liu and Xin Liu and Yiyuan Liu and Haoyu Lu and Shanghao Lu and Fuli Luo and Shirong Ma and Xiaotao Nie and Tian Pei and Yishi Piao and Junjie Qiu and Hui Qu and Tongzheng Ren and Zehui Ren and Chong Ruan and Zhangli Sha and Zhihong Shao and Junxiao Song and Xuecheng Su and Jingxiang Sun and Yaofeng Sun and Minghui Tang and Bingxuan Wang and Peiyi Wang and Shiyu Wang and Yaohui Wang and Yongji Wang and Tong Wu and Y. Wu and Xin Xie and Zhenda Xie and Ziwei Xie and Yiliang Xiong and Hanwei Xu and R. X. Xu and Yanhong Xu and Dejian Yang and Yuxiang You and Shuiping Yu and Xingkai Yu and B. Zhang and Haowei Zhang and Lecong Zhang and Liyue Zhang and Mingchuan Zhang and Minghua Zhang and Wentao Zhang and Yichao Zhang and Chenggang Zhao and Yao Zhao and Shangyan Zhou and Shunfeng Zhou and Qihao Zhu and Yuheng Zou},
      year={2024},
      eprint={2401.02954},
      archivePrefix={arXiv},
      primaryClass={cs.CL},
      url={https://arxiv.org/abs/2401.02954}, 
}

@article{MACAGNO202267,
title = {Argumentation profiles and the manipulation of common ground. The arguments of populist leaders on Twitter},
journal = {Journal of Pragmatics},
volume = {191},
pages = {67-82},
year = {2022},
issn = {0378-2166},
doi = {https://doi.org/10.1016/j.pragma.2022.01.022},
url = {https://www.sciencedirect.com/science/article/pii/S0378216622000285},
author = {Fabrizio Macagno},
}

@article{Egelhofer,
    author = {Egelhofer, Jana Laura and Boyer, Ming and Lecheler, Sophie and Aaldering, Loes},
    title = {Populist attitudes and politicians’ disinformation accusations: effects on perceptions of media and politicians},
    journal = {Journal of Communication},
    volume = {72},
    number = {6},
    pages = {619-632},
    year = {2022},
    month = {10},
    issn = {0021-9916},
    doi = {10.1093/joc/jqac031},
    url = {https://doi.org/10.1093/joc/jqac031},
    eprint = {https://academic.oup.com/joc/article-pdf/72/6/619/47265750/jqac031.pdf},
}

@article{RODRIGOGINES,
title = {A systematic review on media bias detection: What is media bias, how it is expressed, and how to detect it},
journal = {Expert Systems with Applications},
volume = {237},
pages = {121641},
year = {2024},
issn = {0957-4174},
doi = {https://doi.org/10.1016/j.eswa.2023.121641},
url = {https://www.sciencedirect.com/science/article/pii/S0957417423021437},
author = {Francisco-Javier Rodrigo-Ginés and Jorge Carrillo-de-Albornoz and Laura Plaza},
}

@article{Mina_Momeni,
author = {Mina Momeni},
title ={Artificial Intelligence and Political Deepfakes: Shaping Citizen Perceptions Through Misinformation},
journal = {Journal of Creative Communications},
volume = {20},
number = {1},
pages = {41-56},
year = {2025},
doi = {10.1177/09732586241277335},
URL = {https://doi.org/10.1177/09732586241277335},
eprint ={https://doi.org/10.1177/09732586241277335},

}

@misc{APP,
    title={The American Presidency Project},
    author={John Woolley and Gerhard Peters},
    year={2025},
    howpublished = {\url{https://www.presidency.ucsb.edu}},
    note = {Accessed: 14 November 2025}
}

\end{document}